\documentclass[11pt]{article}

\usepackage[final]{acl}

\usepackage{times}
\usepackage{latexsym}
\usepackage{amsmath,amsfonts,bm}
\usepackage{hyperref}
\usepackage{url}
\usepackage{graphicx}
\usepackage{booktabs}
\usepackage{tabularx}

\usepackage[most]{tcolorbox}
\usepackage{graphicx} 
\usepackage{lipsum}

\tcbset{
  colframe=black!70,
  colback=white,
  boxsep=2pt,
  arc=3pt,
  outer arc=3pt,
  fonttitle=\bfseries,
  title filled
}

\usepackage{xcolor} 
\usepackage{multirow}

\usepackage{needspace}
\usepackage[framemethod=default]{mdframed}
\usepackage{xcolor}
\usepackage{enumitem}
\usepackage{float}

\newfloat{prompt}{htbp}{lop}
\floatname{prompt}{Prompt}

\newmdenv[
    linecolor=black!75,
    backgroundcolor=black!5!white,
    linewidth=1pt,
    innertopmargin=7pt,   
    innerbottommargin=7pt,
    innerleftmargin=5pt,
    innerrightmargin=5pt,
]{promptboxstyle}

\usepackage[T1]{fontenc}

\usepackage[utf8]{inputenc}

\usepackage{microtype}

\usepackage{inconsolata}

\usepackage{graphicx}

\title{Reasoning Beyond Literal: Cross-style Multimodal Reasoning for Figurative Language Understanding}

\author{
Seyyed Saeid Cheshmi\hspace{0.7em}
Hahnemann Ortiz\hspace{0.7em}
James Mooney\hspace{0.7em}
Dongyeop Kang\hspace{0.7em} \\
University of Minnesota  \\
\texttt{\{chesh014, orit0001, moone174, dongyeop\}@umn.edu}
}

\begin{document}
\maketitle
\begin{abstract}

Vision–language models (VLMs) have demonstrated strong reasoning abilities in literal multimodal tasks such as visual mathematics and science question answering. However, figurative language—such as sarcasm, humor, and metaphor—remains a significant challenge, as it conveys intent and emotion through subtle incongruities between expressed and intended meanings. In multimodal settings, accompanying images can amplify or invert textual meaning, demanding models that reason across modalities and account for subjectivity.
We propose a three-step framework for developing \textit{efficient multimodal reasoning models} that can (i) interpret multimodal figurative language, (ii) provide transparent reasoning traces, and (iii) generalize across multiple figurative styles. 
Experiments across four styles show that (1) incorporating reasoning traces substantially improves multimodal figurative understanding, (2) reasoning learned in one style can transfer to others—especially between related styles like sarcasm and humor, and (3) training jointly across styles yields a generalized reasoning VLM that outperforms much larger open- and closed-source models.
Our findings show that lightweight VLMs with verifiable reasoning achieve robust cross-style generalization while providing inspectable reasoning traces for multimodal tasks. The code and implementation are available at \url{https://github.com/scheshmi/CrossStyle-MMR}.
\end{abstract}

\section{Introduction}
Human communication often utilizes expressions that convey meanings beyond their literal interpretation.
Figurative language, including sarcasm, humor, metaphor, and offense, is essential in expressing intent, emotion, and perspective. 
Understanding such expressions is vital for sentiment analysis, social media moderation, and human–AI interaction \citep{yang2022multimodal, muse2024}.
However, figurative communication often depends on subtle incongruities between what is said and what is meant \citep{camp2020sarcasm}, making it particularly difficult for language models \citep{jang2025difficult}.

The challenge becomes even more complex in multimodal contexts, where text is paired with images, videos, or other signals. In such cases, visual cues can amplify or invert textual meaning \citep{wang2025can}, requiring models to detect semantic discrepancies across modalities and to reason about subjective interpretations \citep{zhou2025semirnet}.
Despite rapid progress in vision–language models (VLMs), current systems remain limited in handling these nuanced multimodal figurative phenomena \citep{bojic2025comparing}.

\begin{figure}
    \centering
    \includegraphics[width=1.05\linewidth]{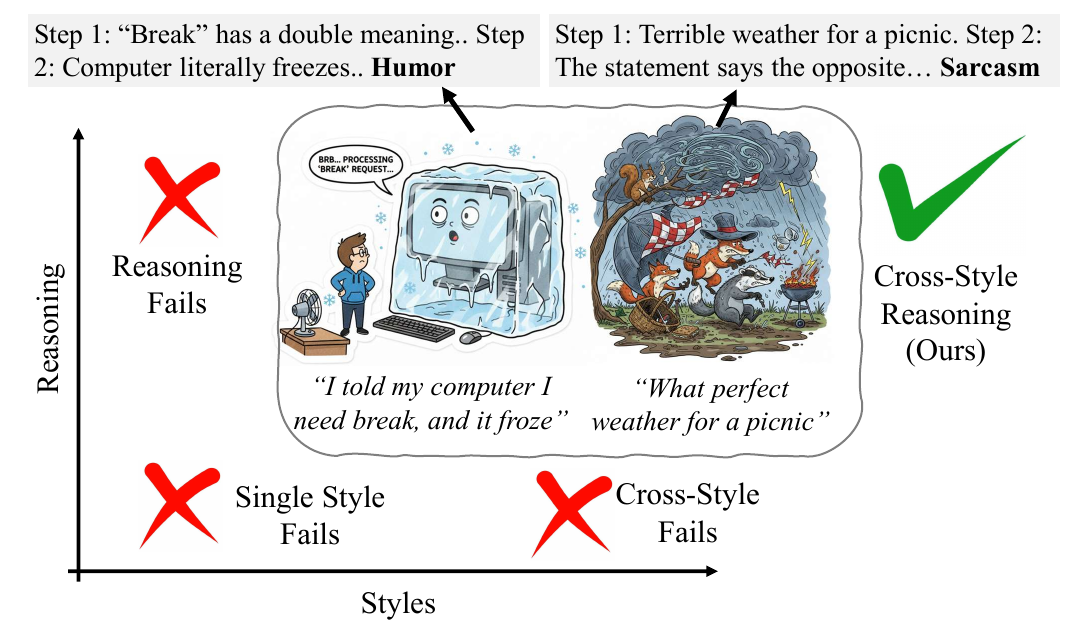}
    \caption{Previous work has focused on single styles and has not explored multimodal figurative language understanding. Our study examines whether incorporating reasoning can improve multimodal figurative understanding, and further, whether it can enable the development of a reasoning model capable of understanding multiple styles. }
    \label{fig:concept}
\end{figure}

Recent advances in reasoning language models have shown that chain-of-thought (CoT) prompting can substantially improve interpretability and performance across analytical domains such as mathematics and science \citep{lu2023mathvista, lu2022learn, kembhavi2016diagram}. Yet, whether such reasoning mechanisms can enhance multimodal figurative language understanding remains unexplored. Developing this capability requires datasets containing explicit reasoning traces—resources that are notably scarce for figurative language.

To bridge this gap, we introduce a three-step reasoning framework for multimodal figurative understanding. Building on \citet{ho2022large,zhang2024improve}, we first distill CoT reasoning paths from a large open-source teacher model \citep{grattafiori2024llama}. We then enhance a lightweight student model through Supervised Fine-Tuning (SFT) on these distilled traces, followed by Reinforcement Learning with Verifiable Rewards (RLVR). This pipeline enables small VLMs to internalize structured reasoning and generate interpretable predictions for figurative multimodal tasks.

Extensive experiments across four figurative styles—sarcasm, humor, offense, and metaphor—demonstrate that:
(1) Incorporating CoT reasoning markedly improves multimodal figurative understanding.
(2) Reasoning knowledge learned in one style can transfer to related styles, particularly between sarcasm and humor.
(3) Joint training across all styles produces a generalized reasoning VLM that outperforms much larger open- and closed-source models (e.g., Gemini 2.5 Flash, LLaMA-90B-Vision-Instruct) on most benchmarks.

To our knowledge, this is the first work to induce structured cross-style reasoning in VLMs using RLVR. Our approach establishes a lightweight but powerful framework for generalizable multimodal figurative language understanding with transparent decision paths.



\section{Related work}
\subsection{Figurative Language Understanding}
Figurative language has been extensively studied in Natural Language Processing (NLP), with research spanning sarcasm, metaphor, idioms, humor, and hyperbole. 
Early approaches relied on textual cues and incongruity signals, while recent datasets have broadened coverage across multiple styles. 
\citet{chakrabarty2022flute} released FLUTE, a dataset of 9,000 figurative instances (sarcasm, simile, metaphor, idioms), created via a model-in-the-loop approach with GPT-3 and human annotators, enabling progress in figurative language understanding. In \citet{jang-etal-2023-figurative}, authors probed the performance of several models on figurative language classification across sarcasm, similes, idioms, and metaphors, conducting experiments to highlight key differences in model behavior and to analyze the linguistic characteristics of these four figurative types. 
Moreover, \citet{lai2023multilingual} investigated multilingual detection across several figurative styles, highlighting the promise of unified modeling. 

Beyond text-only approaches, recent work has explored figurative language understanding in multimodal settings. \citet{saakyan2023vflute} introduced V-FLUTE, a dataset for visual figurative entailment with explanations across metaphors, idioms, sarcasm, and humor, showing that even strong language models struggle to generalize across styles, particularly in images. Building on V-FLUTE, the figurative language shared task consolidated multi-style datasets and emphasized identifying visual entailment relationships in multimodal instances \citep{kulkarni2024report}. 

Additionally, the Multimodal Sarcasm Explanation (MuSE) task formalized the need for models not only to detect but also to justify figurative intent in image-text pairs \citep{desai2022nice}. Although a few studies have investigated figurative language detection through reasoning \citep{tian-etal-2024-theory, yao2025sarcasm} or explainable annotations or eye movements \cite{hayati2021doesbertlearnhumans,delangis2023comparativestudytextualsaliency}, they focus on a single style (sarcasm or metaphor) and text-only settings and cross-style setting \citep{kang2019style,das2023balancingeffecttrainingdataset} has not yet been explored. In this work, we study multimodal reasoning for figurative language and investigate cross-style knowledge transfer.

\begin{figure*}[t]
    \centering
    \includegraphics[width=\textwidth,trim=0cm 0cm 0cm 0cm,clip]{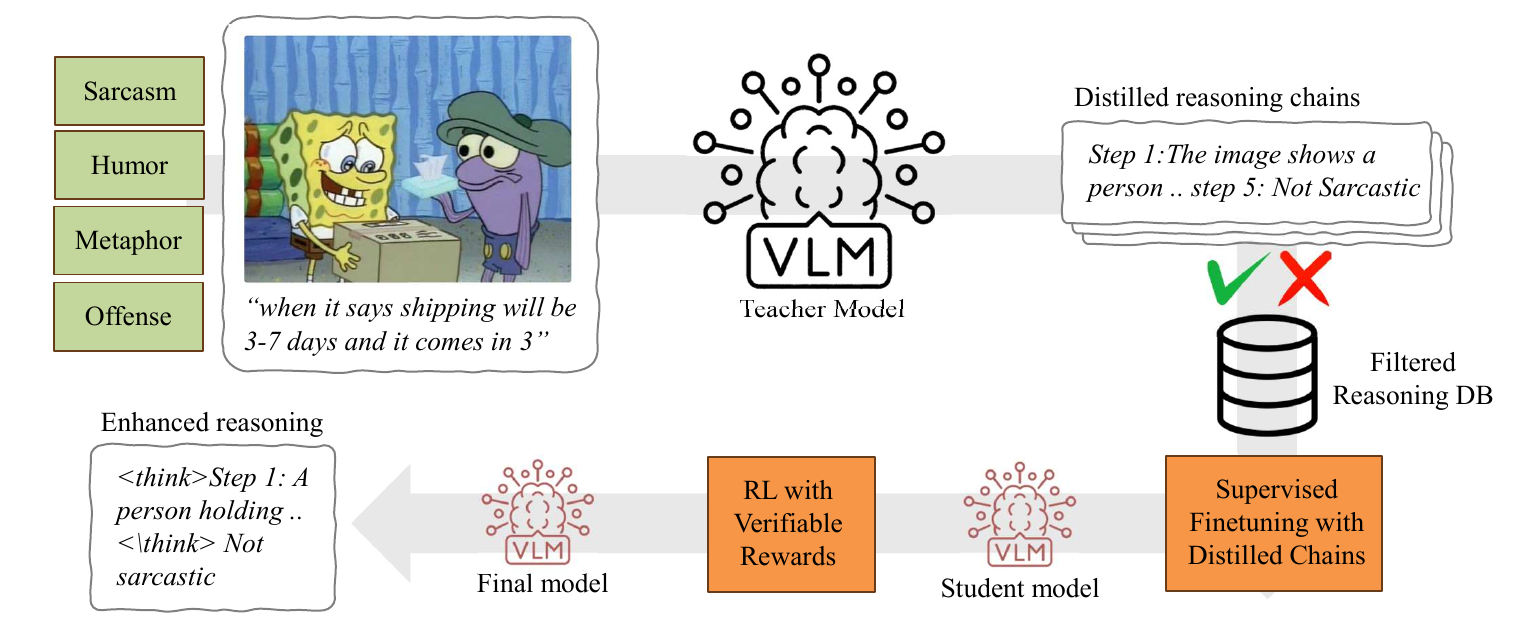}
    \caption{Overall workflow of the proposed method.}
    \label{fig:method}
\end{figure*}

\subsection{Multimodal Reasoning}
The reasoning capabilities of VLMs have been studied primarily in domains such as mathematics \cite{wang2024measuring} and science \cite{lu2022learn, kembhavi2016diagram}. Approaches such as RePIC \cite{oh2025repic} and VLM-R1 \cite{shen2025vlm} typically apply reinforcement learning to align a vision–language backbone with verifiable, task-level rewards but do not explicitly model multi-step chain-of-thought reasoning. These methods replace preference models with deterministic validators (e.g., exact-match or BLEU) and optimize the policy using Group Relative Policy Optimization (GRPO) or Proximal Policy Optimization (PPO) variants with a KL regularizer. Other approaches, such as SVQA-R1 \cite{wang2025svqa} and STAR-R1 \cite{li2025star}, employ view-consistent or transformation-invariant objectives for Visual Question Answering (VQA). More recent work such as Visionary-R1 \cite{xia2025visionary}, enforces image interpretation prior to reasoning to mitigate shortcut reliance and improve generalization. However, these reasoning frameworks have primarily been validated on objective tasks like mathematics and science, where ground truth is objectively verifiable. It remains an open question whether such verifiable reasoning pipelines can be effectively adapted to the subjective and nuanced domain of figurative language understanding.
\section{Method}

Our goal is to equip compact VLMs with explicit multimodal reasoning capabilities that enable them to interpret and generalize across different figurative styles.

Building on recent findings that large models can serve as reasoning teachers \citep{ho2022large} and that reinforcement learning with verifiable rewards (RLVR) can enhance self-evolved reasoning in language models \citep{wen2025reinforcement, guo2025deepseek}, we propose a three-stage reasoning framework illustrated in Figure \ref{fig:method}. 
The approach consists of:
(1) Chain-of-Thought Distillation from a large teacher VLM to extract structured reasoning traces;
(2) Supervised Fine-Tuning on a smaller model to internalize these reasoning patterns; and 
(3) RLVR-based optimization using verifiable rewards from ground-truth labels to further refine reasoning consistency and factual correctness.
Importantly, this framework can be applied both per style and in a combined training regime, the latter producing our best-performing, generalized reasoning VLM.

\subsection{Step 1: Chain-of-Thought Extraction from the Teacher}
We first employ a large VLM as a reasoning teacher to generate structured CoT explanations for multiple figurative styles (sarcasm, humor, offense, and metaphor) under zero-shot prompting. Each CoT follows a standardized five-step schema: image description, caption interpretation, mismatch detection, intent inference, and final label prediction. Table \ref{tab:cot_examples} provides some examples of generated CoTs for different styles.

Generated samples are automatically filtered to remove malformed or incorrect reasoning paths, retaining only valid CoTs for downstream use. This step yields a high-quality reasoning corpus that captures multimodal cues, pragmatic intent, and contextual inconsistencies—knowledge rarely available in existing datasets. Prompts and filtering rules are provided in Appendix \ref{sec:appendix_a1}.

\subsection{Step 2: Chain-of-Thought distillation via Supervised Fine Tuning}
Next, we fine-tune a smaller student model on the distilled reasoning corpus. This SFT stage acts as a reasoning warm-up, teaching the model to emulate the teacher’s explicit CoT style and to produce interpretable intermediate steps.
Through this distillation, the smaller VLM acquires a foundation for step-wise multimodal reasoning while preserving computational efficiency. The SFT model serves as the reference policy for subsequent RLVR training, ensuring stable optimization and alignment between reasoning structure and prediction accuracy.

\subsection{Step 3: Reinforcement Learning with Verifiable Rewards }
Finally, we refine the SFT model using RLVR implemented via GRPO algorithm \citep{guo2025deepseek}. Unlike Reinforcement Learning from Human Feedback (RLHF), RLVR replaces human preference models with deterministic verification functions, enabling scalable, objective evaluation of prediction accuracy and structural adherence.

Given an input instance $q$, the model’s output $o$—comprising both a reasoning trace and a final answer—is evaluated using a binary verifiable reward function $R(q,o)$ that measures factual correctness and adherence to the required reasoning format:
\[
R = R_{\text{acc}} + R_{\text{format}}
\]
where
\begin{equation}
R_{\text{acc}} =
\begin{cases}
1, & \text{if prediction matches ground truth}, \\
0, & \text{otherwise}.
\end{cases}
\end{equation}


The RL objective is to maximize the expected verifiable reward while maintaining proximity to the SFT reference policy through KL-regularized optimization:
\begin{equation}
\max_{\pi_\theta} \; 
\mathbb{E}_{o\sim\pi_\theta(q)}[R(q,o)] 
- \beta \, \text{KL}\!\left[\pi_\theta(o|q)\,||\,\pi_{\text{ref}}(o|q)\right].
\end{equation}

This process reinforces accurate and well-structured reasoning traces without relying on subjective human labels.

The model’s output is expected to include two parts: a reasoning process, enclosed within \texttt{<think></think>} tags, which explains how the model integrates visual and textual cues to arrive at its prediction, and a final prediction, enclosed within \texttt{<answer></answer>} tags. If the output satisfies these formatting constraints, the format reward is assigned a value of 1; otherwise, it is set to 0. The prompt used here can be found in the Appendix \ref{sec:appendix_a2}.

\subsection{Unified Training for Generalized Figurative Reasoning}
While the three stages can be executed independently for each figurative style, our \textbf{combined training strategy}, performing SFT + RLVR jointly on all styles, achieves the best overall performance. This unified optimization enables \textbf{cross-style knowledge transfer}, allowing the model to leverage shared reasoning patterns across different styles.





\section{Experimental Setup}



\subsection{Style Selection and Datasets}
We focus on four figurative language styles—\textit{sarcasm}, \textit{humor}, \textit{offense}, and \textit{metaphor}.  
We leverage three multimodal datasets covering diverse figurative styles. Each dataset provides paired textual and visual information, enabling analysis of cross-modal incongruities central to figurative expression.  

\paragraph{MMSD2.0 \citep{qin-etal-2023-mmsd2}.}  
The Multi-Modal Sarcasm Detection 2.0 benchmark contains roughly 25{,}000 image--text pairs annotated for sarcasm. It offers rich context for studying multimodal incongruity and intent recognition.  

\paragraph{Memotion \citep{chhavi2020memotion}.}  
This SemEval-2020 Task 8 dataset comprises about 9{,}800 memes annotated for multiple emotional dimensions, including \textit{humor} and \textit{offense}. Its varied labels and human-annotated judgments make it ideal for evaluating subjectivity and overlapping stylistic cues.  

\paragraph{MultiMET \citep{zhang-etal-2021-multimet}.}  
MultiMET includes 10{,}437 image--text pairs annotated for \textit{metaphor} detection. It distinguishes text-dominant, image-dominant, and complementary metaphors, supporting analysis of abstract, cross-modal reasoning.  

For all datasets, we adopt standard splits: the default training and test sets for MMSD2.0 and Memotion, and an 80/20 split for MultiMET. CoT reasoning traces are distilled exclusively from the training data of each dataset.

\subsection{Models}
We employ LLaMA3.2-90B-Vision-Instruct \cite{grattafiori2024llama} as the reasoning teacher model for the first step of our approach, and use Qwen2.5-VL-3B-Instruct \cite{bai2025qwen2} as the student model in the second and third steps.
All experiments are formulated as binary classification tasks, where the goal is to determine whether an image--text pair expresses a specific figurative meaning. Our primary model backbone is \texttt{Qwen2.5-VL-3B-Instruct}, a compact yet capable VLM chosen to evaluate reasoning scalability under constrained model size. 
All model evaluations and training are performed with \textbf{zero-shot prompting}, ensuring fair and consistent comparisons across methods.  

We evaluate five experimental configurations to assess the contribution of each training stage:  
(1) \texttt{Zero-shot CoT Prompting},  
(2) \texttt{SFT on Binary Labels},  
(3) \texttt{SFT on Distilled CoTs},  
(4) \texttt{GRPO-only (RLVR without SFT)}, and  
(5) \texttt{SFT $\rightarrow$ GRPO (Full Pipeline)}.  
Table \ref{tab:configs} includes detailed configurations.

\begin{table}[t]
\centering
\caption{Training model configurations.}
\label{tab:configs}
\setlength{\tabcolsep}{5pt}
\begin{tabular}{lp{5cm}}
\toprule
\textbf{Setup} & \textbf{Description} \\
\midrule
Zero-shot CoT & Prompt teacher/student to produce CoT without gradient updates \\
SFT-Binary & Supervised fine-tuning on gold binary labels only \\
SFT-CoT & Supervised fine-tuning on distilled teacher CoTs \\
GRPO-only & RLVR via GRPO starting from base model (no SFT) \\
SFT$\rightarrow$GRPO & SFT-CoT warm-up followed by RLVR via GRPO (full pipeline) \\
\bottomrule
\end{tabular}
\end{table}

\subsection{Implementation Details}
SFT is performed for 5 epochs with a learning rate of 2e--4 using cosine annealing. RLVR via GRPO is conducted for 2 epochs with a learning rate of 1e--5 and 8 rollouts per sample.  

All experiments are executed on two NVIDIA H100 GPUs. The SFT stage equips the model with baseline reasoning capabilities, while RLVR refines its verifiable reasoning and alignment between reasoning traces and final predictions. Together, these stages enable the development of a compact, interpretable, and generalized multimodal reasoning VLM.



\begin{table*}[t]
\centering
\caption{Pipeline ablation: accuracy (Acc) and F1 across four figurative styles. Best score in each subcolumn is \textbf{bold}. All runs use zero-shot prompting.}
\label{tab:pipeline_combined}
\setlength{\tabcolsep}{7pt}
\renewcommand{\arraystretch}{1.12}
\begin{tabular}{l*{4}{cc}}
\toprule
\multirow{2}{*}{\textbf{Experiment}} 
& \multicolumn{2}{c}{\textbf{Sarcasm}} 
& \multicolumn{2}{c}{\textbf{Humor}} 
& \multicolumn{2}{c}{\textbf{Offense}} 
& \multicolumn{2}{c}{\textbf{Metaphor}} \\
\cmidrule(lr){2-3}\cmidrule(lr){4-5}\cmidrule(lr){6-7}\cmidrule(lr){8-9}
& \textbf{Acc} & \textbf{F1} & \textbf{Acc} & \textbf{F1} & \textbf{Acc} & \textbf{F1} & \textbf{Acc} & \textbf{F1} \\
\midrule
Zero-shot CoT Prompting   & 63.88 & 0.41 & 49.82 & 0.68 & 46.72 & 0.44 & 53.89 & 0.63 \\
SFT on Binary Labels      & 82.02 & 0.79 & 66.29 & 0.74 & 50.57 & 0.62 & 59.67 & 0.69 \\
SFT on Distilled CoTs     & 83.91 & 0.77 & 70.73 & 0.78 & 52.15 & 0.59 & 62.19 & 0.73 \\
GRPO-only                 & 72.19 & 0.43 & 65.85 & 0.74 & 48.33 & 0.49 & 59.21 & 0.66 \\
\textbf{SFT $\rightarrow$ GRPO} 
                          & \textbf{86.82} & \textbf{0.81} & \textbf{78.91} & \textbf{0.89} & \textbf{59.51} & \textbf{0.75} & \textbf{69.24} & \textbf{0.80} \\
\bottomrule
\end{tabular}
\end{table*}

\section{Results}
We evaluate our approach on three questions:  
\textit{(1) Does CoT reasoning improve multimodal figurative understanding?}  
\textit{(2) Can reasoning learned for one figurative style transfer to others?}  
\textit{(3) Can a single generalized reasoning VLM match or surpass much larger models?}

\subsection{Does CoT reasoning enhance multimodal figurative understanding?}
\paragraph{Observation.} 
Table \ref{tab:pipeline_combined} present the accuracy and F1 score results for Qwen2.5-VL-3B-Instruct across four different styles under the following setups: zero-shot CoT prompting, SFT on binary labels, SFT on distilled CoT, GRPO-only, and SFT + GRPO on the same style. 
As shown, fine-tuning the model on binary labels outperforms zero-shot CoT prompting across all styles, which aligns with previous findings \cite{bojic2025comparing} highlighting the challenges of figurative language understanding for LLMs and VLMs. Furthermore, performing SFT on distilled CoTs to enable reasoning capabilities yields notable gains: \(\sim1.9\%\) accuracy improvement on sarcasm detection, \(\sim4.5\%\) on humor detection, \(\sim2.4\%\) on offense detection, and \(\sim2.6\%\) on metaphor detection. Building on this, applying RLVR with GRPO on top of the SFT models leads to even greater improvements \(\sim3\%\) for sarcasm detection, \(\sim8\%\) for humor detection, \(\sim7\%\) for offense detection, and \(\sim7\%\) for metaphor detection. These results demonstrate that RLVR effectively enhances CoT reasoning when the model is first equipped with initial reasoning capabilities through SFT on distilled CoTs. Additionally, results from the GRPO-only setup show that while some improvement over zero-shot CoT prompting is possible, the absence of initial CoT reasoning and the inherent complexity of figurative language highlight the critical role of SFT on CoTs as a warm-up step.


\subsection{Can CoT learned in one style transfer to other styles?}
We test cross-style transfer by performing SFT-CoT on a source style and RLVR on a distinct target style. 
Here, we conduct cross-style experiments where we first perform distillation SFT on one style and then apply GRPO on a different style. We repeat this process for all style combinations. 

\begin{figure}[t]
    \centering
    \includegraphics[width=0.98\linewidth]{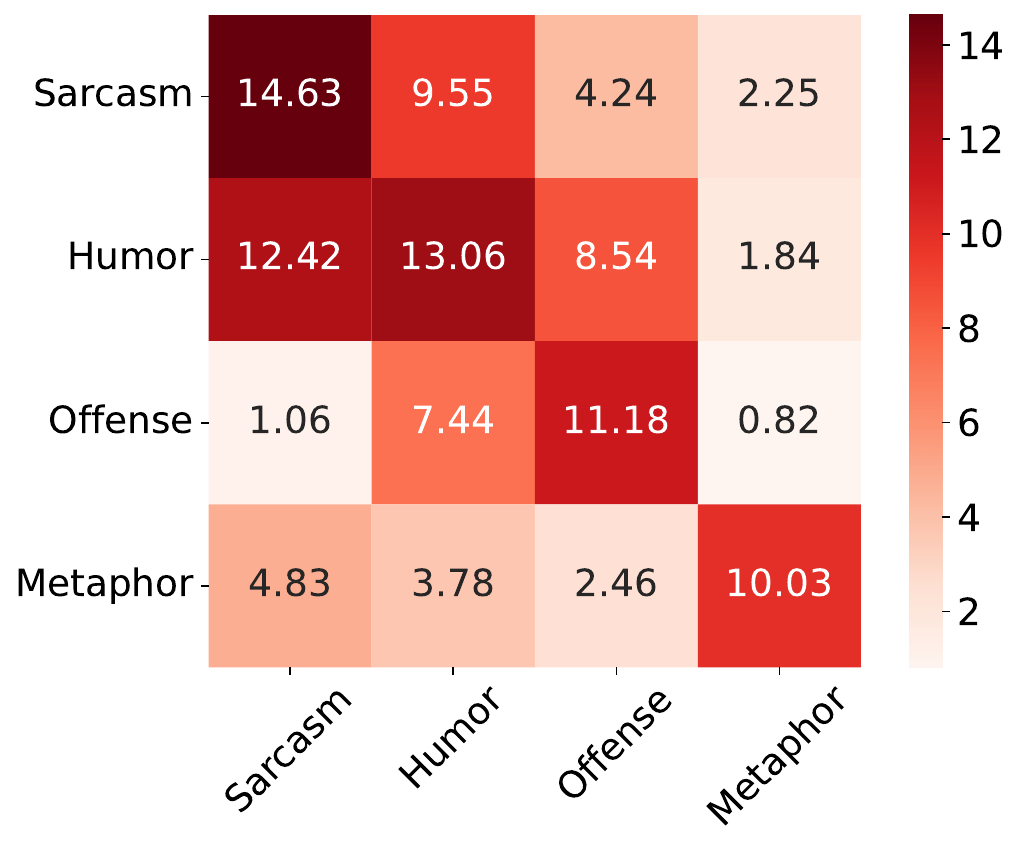}
    \caption{Cross-style gains over GRPO-only. Rows: SFT-CoT source style; columns: RLVR target style. Diagonal cells (same-style) are highest; sarcasm$\leftrightarrow$humor shows the strongest transfer}
    \label{fig:heatmap}
\end{figure}

\paragraph{Observation.} 
Table \ref{tab:combined_transfer} presents the results of these experiments. 
Transfer is strongest between semantically related styles (sarcasm$\leftrightarrow$humor). 
For example, performing SFT on distilled sarcasm CoTs followed by RLVR on humor data yields a \(\sim10\%\) accuracy improvement compared to performing GRPO on the base model.
Figure~\ref{fig:heatmap} visualizes absolute gains over GRPO-only; the diagonal (same-style SFT$\rightarrow$GRPO) is highest, while off-diagonal improvements confirm cross-style knowledge sharing. Metaphor shows weaker transfer into other styles, consistent with its greater semantic distinctiveness \cite{skalicky-crossley-2018-linguistic}. 
These findings suggest that RLVR on top of an SFT model trained on a related style can achieve improved performance without requiring the first two steps of our proposed approach for the target style.

\begin{table*}[t]
\centering
\caption{Cross-style transfer results: accuracy (Acc) and F1 for each target style. Best score per subcolumn is \textbf{bold}.}
\label{tab:combined_transfer}
\setlength{\tabcolsep}{7pt}
\renewcommand{\arraystretch}{1.12}
\begin{tabular}{l*{4}{cc}}
\toprule
\multirow{2}{*}{\textbf{Experiment}} 
& \multicolumn{2}{c}{\textbf{Sarcasm}} 
& \multicolumn{2}{c}{\textbf{Humor}} 
& \multicolumn{2}{c}{\textbf{Offense}} 
& \multicolumn{2}{c}{\textbf{Metaphor}} \\
\cmidrule(lr){2-3}\cmidrule(lr){4-5}\cmidrule(lr){6-7}\cmidrule(lr){8-9}
& \textbf{Acc} & \textbf{F1} & \textbf{Acc} & \textbf{F1} & \textbf{Acc} & \textbf{F1} & \textbf{Acc} & \textbf{F1} \\
\midrule
GRPO-only                         & 72.19 & 0.43 & 65.85 & 0.74 & 48.33 & 0.49 & 59.21 & 0.66 \\
SFT Humor $\rightarrow$ GRPO      & 84.61 & 0.78 & \textbf{78.91} & \textbf{0.89} & 56.87 & 0.67 & 61.05 & 0.72 \\
SFT Sarcasm $\rightarrow$ GRPO    & \textbf{86.82} & \textbf{0.81} & 75.40 & 0.82 & 52.57 & 0.58 & 61.46 & 0.72 \\
SFT Metaphor $\rightarrow$ GRPO   & 77.02 & 0.63 & 69.63 & 0.74 & 50.79 & 0.53 & \textbf{69.24} & \textbf{0.80} \\
SFT Offense $\rightarrow$ GRPO    & 73.25 & 0.57 & 73.29 & 0.80 & \textbf{59.51} & \textbf{0.75} & 60.03 & 0.64 \\
\bottomrule
\end{tabular}
\end{table*}

\begin{table*}[t]
\centering
\caption{Generalized model vs.\ large VLMs: accuracy (Acc) and F1 per style. Best in each subcolumn is \textbf{bold}. All results use zero-shot prompting.}
\label{tab:gen_all_combined}
\renewcommand{\arraystretch}{1.0}
\resizebox{\textwidth}{!}{
\begin{tabular}{l*{4}{cc}}
\toprule
\multirow{2}{*}{\textbf{Model / Setup}} 
& \multicolumn{2}{c}{\textbf{Sarcasm}} 
& \multicolumn{2}{c}{\textbf{Humor}} 
& \multicolumn{2}{c}{\textbf{Offense}} 
& \multicolumn{2}{c}{\textbf{Metaphor}} \\
\cmidrule(lr){2-3}\cmidrule(lr){4-5}\cmidrule(lr){6-7}\cmidrule(lr){8-9}
& \textbf{Acc} & \textbf{F1} & \textbf{Acc} & \textbf{F1} & \textbf{Acc} & \textbf{F1} & \textbf{Acc} & \textbf{F1} \\
\midrule
Gemini 2.5 Flash                                     & 89.34 & 0.84 & 79.13 & 0.89 & 60.95 & 0.77 & \textbf{73.19} & 0.76 \\
Qwen2.5-VL 32B Instruct                                 & 76.78 & 0.64 & 77.14 & 0.87 & 49.37 & 0.52 & 61.31 & 0.69 \\
LLaMA 90B Vision Instruct                            & 88.85 & 0.83 & 78.71 & 0.87 & 55.47 & 0.75 & 67.56 & 0.76 \\
Gemma 3 27B                                          & 54.63 & 0.61 & 62.23 & 0.68 & 45.33 & 0.57 & 58.44 & 0.64 \\
Phi-4 Multimodal                                     & 68.28 & 0.56 & 70.33 & 0.73 & 48.74 & 0.64 & 55.71 & 0.55 \\
LLaVA 1.5 7B                                         & 59.09 & 0.48 & 61.54 & 0.61 & 43.22 & 0.52 & 53.68 & 0.56 \\
\hline
SFT on Combined $\rightarrow$ GRPO (Style specific)  & 87.76 & 0.82 & 78.59 & 0.91 & 60.27 & 0.77 & 68.19 & 0.79 \\
\textbf{SFT on Combined $\rightarrow$ GRPO (Combined)} & \textbf{90.23} & \textbf{0.85} & \textbf{80.71} & \textbf{0.93} & \textbf{61.49} & \textbf{0.79} & 69.14 & \textbf{0.80} \\
\bottomrule
\end{tabular}
}
\end{table*}

\subsection{Can we build a single generalized model across styles?}
We investigate whether our proposed approach can produce a generalized reasoning small VLM capable of understanding all figurative styles (in this case, four styles).
We train a unified model by distilling CoTs from the union of all training data and then running SFT$\rightarrow$GRPO on either (i) each target style or (ii) the combined dataset. We compare against much larger open- and closed-source systems under zero-shot prompting.

\paragraph{Observation.} 
Joint SFT$\rightarrow$GRPO on the combined dataset yields the best overall small-model performance (Table~\ref{tab:gen_all_combined}), surpassing much larger models on sarcasm, humor, and offense, while trailing Gemini 2.5 Flash only on metaphor. This validates that \emph{a single lightweight reasoning VLM} can generalize across styles and remain competitive with far larger systems.

\subsection{Effect of Data Scale vs. Cross-Style Diversity}
\label{sec:fixed_budget}

To isolate the impact of cross-style diversity from training data scale, we conduct a controlled experiment with a fixed training budget of 5k samples. In the \emph{style-specific} setting, separate SFT$\rightarrow$GRPO models are trained using 5k samples from a single style. In the \emph{combined} setting, a unified model is trained on the same total number of samples, uniformly sampled across all four styles (1.25k per style). Evaluation is performed on the original test sets.

As shown in Table \ref{tab:diversity_ablation}, the combined model achieves clear improvements on sarcasm and humor, despite having fewer per-style training examples. This indicates that the gains observed in Section~5.3 are not solely attributable to increased training data size, but also to shared reasoning learned across related figurative styles. In contrast, offense and metaphor exhibit slight performance degradation under unified training, suggesting weaker cross-style transfer for these categories when the training budget is constrained. Overall, these results show that cross-style diversity does not benefit all categories equally, with the most consistent improvements observed for sarcasm and humor.

\begin{table}[h]
\centering
\caption{Comparison of Style-specific vs. Combined models with a fixed training budget (5k samples).}
\label{tab:diversity_ablation}
\begin{tabular}{lcccc}
\toprule
\multirow{2}{*}{\textbf{Style}} & \multicolumn{2}{c}{\textbf{Style specific}} & \multicolumn{2}{c}{\textbf{Combined}} \\
\cmidrule(lr){2-3} \cmidrule(lr){4-5}
& \textbf{Acc} & \textbf{F1} & \textbf{Acc} & \textbf{F1} \\
\midrule
Sarcasm & 80.25 & 0.78 & \textbf{83.91} & \textbf{0.81} \\
Humor & 78.13 & 0.88 & \textbf{79.95} & \textbf{0.89} \\
Offense & \textbf{56.36} & \textbf{0.73} & 55.82 & 0.71 \\
Metaphor & \textbf{66.44} & \textbf{0.77} & 63.15 & 0.72 \\
\bottomrule
\end{tabular}
\end{table}

\subsection{Disagreement Analysis}
To better understand the practical advantage of our approach over general-purpose models, we conducted a systematic disagreement analysis between our best model (SFT $\rightarrow$ GRPO on combined dataset) and Gemini 2.5 Flash. We isolated instances in the test sets where the two models produced conflicting predictions and calculated which model matched the ground truth in these contested cases.

As shown in Table \ref{tab:disagreement}, the models disagreed on a significant portion of the test set. In the sarcasm style , the models disagreed on 529 out of 2,390 samples. Within this disagreement subset, our model correctly predicted 309 samples ($58.4\%$), whereas Gemini was correct in only 220 cases. This indicates that our specialized reasoning framework aligns better with human labels in ambiguous sarcasm instances.

We observed similar trends for humor and offense. In the humor style, our model demonstrated its strongest relative performance, winning $70.0\%$ of the disagreement cases (21 vs. 9). For offense, our model maintained a lead with a $55.7\%$ win rate. However, consistent with the main accuracy results, Gemini 2.5 Flash retained an advantage in the metaphor style, correctly classifying 350 of the 561 contested samples. These results highlight that our lightweight model effectively corrects errors made by larger VLMs in affect-heavy styles like sarcasm and humor, though larger models may still hold an edge in abstract semantic associations required for metaphor.

\begin{table*}[t]
\centering
\caption{Comparison of correctly classified samples when our model and Gemini 2.5 Flash differ.}
\label{tab:disagreement}
\begin{tabular}{lcccc}
\toprule
\textbf{Style} & \textbf{Disagreed} & \textbf{Ours Correct} & \textbf{Gemini Correct} & \textbf{Win Rate (\%)} \\
\midrule
Sarcasm  & 529 & \textbf{309} & 220 & 58.4\% \\
Humor    & 30  & \textbf{21}  & 9   & 70.0\% \\
Offense  & 244 & \textbf{136} & 108 & 55.7\% \\
Metaphor & 561 & 211 & \textbf{350} & 37.6\% \\
\bottomrule
\end{tabular}
\end{table*}

\section{Qualitative Examples Analysis}
In this section, we analyze a few examples where our best model (SFT $\rightarrow$ GRPO on the combined dataset) makes correct predictions, while Gemini 2.5 Flash fails to predict the final label correctly. We focus only on samples for sarcasm style from the MMSD2.0 test set.

\begin{tcolorbox}[title=Example 1:,
    breakable,
    colframe=yellow!60!black,   
    colback=yellow!5,          
    coltitle=white,             
    colbacktitle=yellow!60!black, 
    center title
]
    \begin{center}
        \includegraphics[width=0.5\linewidth]{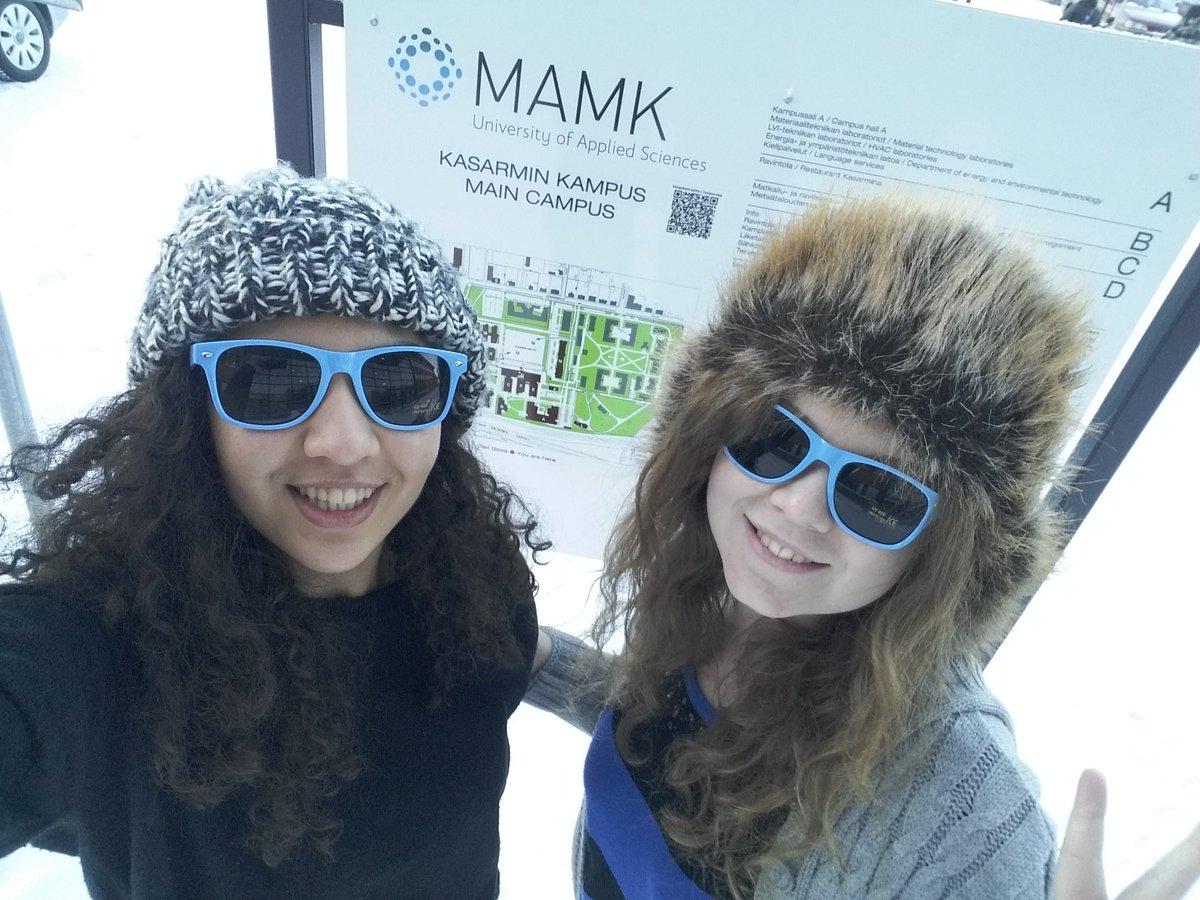} 
    \end{center}
    \textbf{Text:} mikkeli university of applied sciences . it 's too sunny . \# mamk \# finland \# finnips \# uas \# suomi \\
    \textbf{Ground Truth:} Sarcastic \\
    \textbf{Our model CoT:}  <think>Step 1:The image shows a person wearing sunglasses and a hat, standing in front .... Step 2: The caption reads: mikkeli university of applied sciences.... Step 3: There is a mismatch between the image and the caption...... Step 4: it can be inferred that the intent behind the post is sarcastic. The person ....it.</think><answer>sarcastic</answer> \\
    \textbf{Gemini CoT:} Step 1: The image shows two young women posing for a .... .Step 2: The caption states: mikkeli ... Step 3: The image depicts a snowy, winter environment. The women are dressed in warm winter attire, .... the bright conditions and the use of sunglasses shown in the image. There is no clear mismatch. Step 4:  Given the bright, snowy conditions and the fact that both individuals are wearing sunglasses ..... caption's claim; instead, they are congruent. Step 5: not sarcastic. \\
    \textbf{Our Prediction:} Sarcastic \\
    \textbf{Gemini Prediction:} Not sarcastic

\end{tcolorbox}

\textit{Explanation for example 1}:
Our model recognizes sarcasm by noting the contrast between the cheerful photo and the caption complaining it’s “too sunny,” interpreting it as a lighthearted exaggeration rather than a real complaint. However, Gemini takes the post literally, reasoning that bright snow could indeed make it “too sunny,” and therefore sees no contradiction. In other word, our model captures the playful irony, while Gemini overanalyzes environmental realism and misses the humorous intent.

\begin{tcolorbox}[title=Example 2:,
    breakable,
    colframe=green!60!black,   
    colback=green!5,          
    coltitle=white,             
    colbacktitle=green!60!black, 
    center title
]
    \begin{center}
        \includegraphics[width=0.5\linewidth]{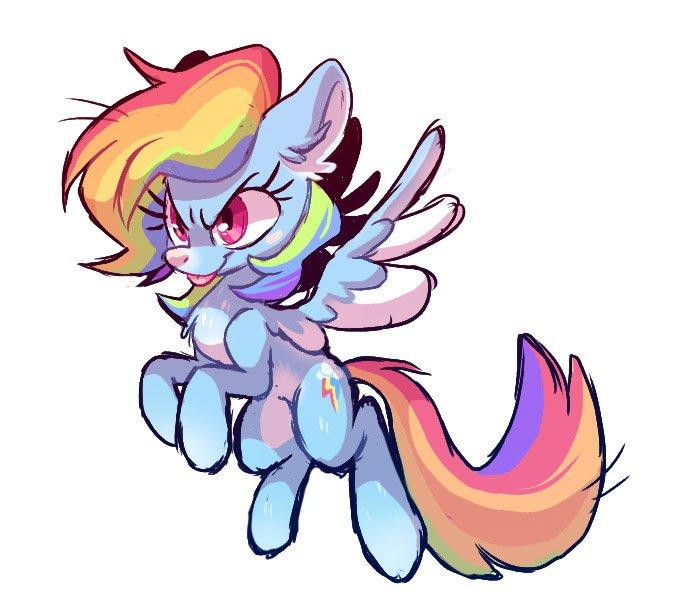} 
    \end{center}
    \textbf{Text:} not sure why i drew rainbuw dersh but i did \\
    \textbf{Ground Truth:} Not sarcastic \\
    \textbf{Our model CoT:} <think>Step 1: The image depicts a cartoon-style ..... Step 2:The caption reads, not sure .... Step 3: . The caption's tone is casual and conversational, suggesting that the .... .Step 4: Based on the congruence between the image and caption, it can be inferred that the intent behind the image is genuine and not sarcastic. .... </think><answer>not sarcastic</answer> \\
    \textbf{Gemini CoT:}  Step 1: The image shows a vibrant, stylized illustration of a light blue pegasus .... Step 2:  The caption says: not sure why ..... Step 3: The image clearly depicts Rainbow Dash, ...... The mismatch lies in the phrase not sure why i drew... Drawing an elaborate piece of fan art typically implies some level of intention, desire, or motivation from the artist......  This is a common form of self-deprecating or ironic humor ....... Step 4:  The artist created a clear, recognizable, and detailed drawing of Rainbow Dash. The caption's statement not sure why i drew rainbuw dersh is likely not meant to be taken literally. Instead, it serves as a humorous, ironic, or self-aware comment. ......... This playful downplaying of the creative act, juxtaposed with the evident effort and clear subject, indicates a sarcastic or ironic intent, rather than genuine confusion. Step 5:  Sarcastic \\
    \textbf{Our Prediction:} Not sarcastic \\
    \textbf{Gemini Prediction:} Sarcastic

\end{tcolorbox}

\textit{Explanation for example 2}:
Our model correctly interprets this pair as not sarcastic, by focusing on the literal meaning of the caption and the cheerful cartoon image. It sees the creator as simply sharing artwork without any hidden irony, recognizing the direct congruence between image and text. Gemini, however, overanalyzes the situation. It interprets the phrase “not sure why I drew rainbuw dersh” as ironic or self-deprecating humor, reasoning that the effort put into the detailed drawing contradicts the claimed uncertainty. This leads Gemini to label the post as sarcastic, even though the creator is genuinely sharing the artwork in a casual, conversational tone.

\paragraph{Observation.} 
Gemini tends to take things too literally or analyze them too much, while our model can pick up on social and figurative signs of sarcasm. In the first example, Gemini focuses on realistic details and logic, so it misses the playful exaggeration that our model correctly sees as sarcastic. In the second example, which isn’t sarcastic, Gemini overthinks a casual phrase like “not sure why I drew,” interpreting it as ironic or self-deprecating. Our model, however, correctly sees that it’s just a simple statement and understands it literally. Therefore, Gemini overanalyzes and misses the subtle, often illogical nature of sarcasm, while our model successfully captures the intended meaning and context.

\section{Conclusion}

In this work, we introduced a three-stage framework that effectively induce reasoning capabilities to compact VLMs for the challenging task of figurative language understanding. Our results demonstrate that explicit reasoning not only enhances figurative comprehension but can also be successfully transferred across related styles. Most notably, our unified model, trained jointly on multiple styles, sets a new benchmark for efficiency, outperforming larger open- and closed-source VLMs while providing transparent reasoning traces that facilitate model inspection. Qualitative analysis further shows that our model captures subtle, non-literal intent that larger models often miss due to overly literal interpretations or excessive contextual over-analysis. The success of our framework has implications beyond figurative language. The core methodology—distilling structured reasoning from a powerful teacher model and refining a smaller student model with verifiable, task-aligned rewards—offers a scalable and computationally efficient paradigm for other complex multimodal tasks. Nevertheless, it is important to recognize the inherent limitations of this approach. The generalization of the student model is fundamentally constrained by the quality and depth of the reasoning traces provided by the teacher. Benefits plateau because the student cannot acquire reasoning patterns that the teacher itself cannot express. Any subtle biases or overlooked nuances in the teacher are inevitably transferred, and potentially amplified, in the student. Future work should investigate how teacher model architectures, sizes, and pre-training domains influence the quality of distilled reasoning to optimize knowledge transfer. While we observed strong cross-style transfer between semantically related styles such as sarcasm and humor, this effect may weaken as the conceptual distance between tasks increases. Moreover, the effectiveness of the RLVR stage relies on the availability of clear, objective ground-truth labels. Exploring more sophisticated reward functions—such as nuanced metrics for reasoning quality or penalties for overconfident incorrect predictions—could further enhance the framework, particularly in sensitive or high-stakes applications.
\vspace{-0.3em}
\section{Limitation}
Although our proposed framework achieves strong performance and outperforms larger models in figurative language understanding, our study has some limitations. First, the experiments should be extended to different models in both the CoT distillation and subsequent training steps to examine the behavior of larger models under various settings. Second, our study lacks out-of-distribution experiments to evaluate whether the CoT reasoning capability learned by VLMs can generalize across different datasets. Third, incorporating additional figurative styles and new datasets would be valuable. Fourth, a comparison with both open-source and closed-source models in a few-shot setting would be helpful. Finally, conducting an ablation study to isolate the effects of CoT filtering and reward formatting is essential.


\bibliography{custom}
\clearpage
\appendix
\label{sec:appendix}
\section{Appendix}
\subsection{CoT Prompt}
\label{sec:appendix_a1}
We use a decomposed CoT prompting strategy, inspired by \citet{khot2022decomposed} and \citet{zhou2022least} to break the problem into four steps. 
At each step, we instruct the model to focus on a specific aspect of the input data.
Prompt \ref{prompt:sarcasm_prompt} detects sarcasm, Prompt \ref{prompt:humor_prompt} humor, Prompt \ref{prompt:offensive_prompt} offense, and Prompt \ref{prompt:metaphor_prompt} metaphor. After generating the CoT, we check, using regex, whether it contains all five steps and whether the final prediction matches the ground truth, discarding any samples that do not meet these criteria. Table \ref{tab:cot_examples} presents examples of generated CoTs.




\begin{prompt}[ht!]
    \centering
    \begin{promptboxstyle}
        You are an expert at detecting sarcasm in images and text. Analyze the provided image and caption to determine if the pair is sarcastic or not sarcastic. Provide your reasoning in the following format:
        \vspace{1ex}
        
        \textbf{Step 1:} What the image shows: [Detailed description of the image content] \\
        \textbf{Step 2:} What the caption says: [Quote or paraphrase the caption] \\
        \textbf{Step 3:} Detecting mismatch: [Explain if there is a mismatch or congruence between the image and caption, and why] \\
        \textbf{Step 4:} Inference of intent: [Conclude whether the intent is sarcastic or not based on the mismatch/congruence] \\
        \textbf{Step 5:}  [Provide your final answer in the form of 'sarcastic' or 'not sarcastic']   
    \end{promptboxstyle}
    \caption{CoT prompt to detect sarcasm.}
    \label{prompt:sarcasm_prompt}
\end{prompt} 



\begin{prompt}[ht!]
    \centering
    \begin{promptboxstyle}
        You are an expert at detecting humor in images and text. When given an image and text, analyze whether the content is humorous or not. Provide your reasoning process in the following format:      
        \vspace{1ex}
        
        \textbf{Step 1:} What the image shows: [Detailed description of the image content] \\
        \textbf{Step 2:} What the caption says: [Quote or paraphrase the caption] \\
        \textbf{Step 3:} Humor cues: [Explain if there are elements such as exaggeration, wordplay, absurdity, or incongruity between the image and caption that make the content humorous] \\
        \textbf{Step 4:} Inference of intent: [Conclude whether the intent is humorous or not based on the cues] \\
        \textbf{Step 5:}  [Provide your final answer in the form of "humorous" or "not humorous"]       
    \end{promptboxstyle}
    \caption{CoT prompt to detect humor.}
    \label{prompt:humor_prompt}
\end{prompt}


\begin{prompt}[ht!]
    \centering
    \begin{promptboxstyle}
        You are an expert at detecting offensive content in images and text. When given an image and text, analyze whether the content is offensive or not. Provide your reasoning process in the following format:     
        \vspace{1ex}
        
        \textbf{Step 1:} What the image shows: [Detailed description of the image content] \\
        \textbf{Step 2:} What the caption says: [Quote or paraphrase the caption] \\
        \textbf{Step 3:} Offense cues: [Explain if there are elements such as hate speech, slurs, derogatory language, demeaning stereotypes, harassment, or explicit insults that make the content offensive] \\
        \textbf{Step 4:} Context and intent: [Discuss whether the content was likely meant to harm, insult, or demean someone, or if it might be interpreted as offensive even without harmful intent] \\
        \textbf{Step 5:}  [Provide your final answer in the form of "offensive" or "not offensive"]       
    \end{promptboxstyle}
    \caption{CoT prompt to detect offense.}
    \label{prompt:offensive_prompt}
\end{prompt}



\begin{prompt}[ht!]
    \centering
    \begin{promptboxstyle}
        You are an expert at detecting metaphors in images and text. When given an image and text, analyze whether the content uses metaphorical language or not. Provide your reasoning process in the following format:       
        \vspace{1ex}
        
        \textbf{Step 1:} What the image shows: [Detailed description of the image content] \\
        \textbf{Step 2:} What the caption says: [Quote or paraphrase the caption] \\
        \textbf{Step 3:} Metaphor cues: [Explain if there are figurative expressions, symbolic comparisons, or non-literal meanings that connect the caption and the image] \\
        \textbf{Step 4:} Interpretation: [Discuss what abstract idea, concept, or meaning the metaphor might be conveying] \\
        \textbf{Step 5:}  [Provide your final answer in the form of "metaphorical" or "not metaphorical"]        
    \end{promptboxstyle}
    \caption{CoT prompt to detect metaphor.}
    \label{prompt:metaphor_prompt}
\end{prompt}

\begin{table}[ht]
\centering
\small
\caption{Generated reasoning chains for four figurative language styles.}
\setlength{\tabcolsep}{4pt}
\renewcommand{\arraystretch}{1.1}
\begin{tabular}{p{1.5cm} p{6cm}}
\toprule
\textbf{Style} & \textbf{Generated Chain-of-Thought} \\
\midrule
\textbf{Sarcasm} &
Step 1: The image shows a map of Malaysia with a green area in the center, which appears to be a lake or a body of water. The map also shows the surrounding landmasses and the coastline. 
Step 2: The caption reads, “this is how our currency is shrinking \#ringgitladesh.” It refers to the Malaysian currency (Ringgit) and implies that it is losing value. 
Step 3: There is a mismatch between the image and the caption. The image shows a map unrelated to currency, while the caption refers to economic value. 
Step 4: The mismatch suggests the post is sarcastic, using irony rather than serious commentary to highlight the shrinking value of the currency. 
Step 5: Sarcastic. \\[4pt]

\textbf{Humor} &
Step 1: The image depicts a man walking down the street with a woman on his arm while looking back at another woman. The man wears a blue plaid shirt; the woman on his arm is in a light blue top; the woman he looks at wears a red dress. 
Step 2: The caption reads “People of the future” above the man’s head, “Traditional human language” above the woman on his arm, and “Communicating entirely through the Distracted Boyfriend meme” above the woman he is looking at. 
Step 3: The humor arises from applying the well-known Distracted Boyfriend meme to a futuristic scenario where people communicate only through memes. The absurd combination creates incongruity and amusement. 
Step 4: The intent is to be humorous by exaggerating modern meme culture. 
Step 5: Humorous. \\[4pt]

\textbf{Offense} &
Step 1: The image shows two side-by-side photos of a man with a shaved head, wearing a tuxedo. The left photo looks natural; the right has exaggerated, cartoonish features. 
Step 2: “Stop doing this to your pics.” 
Step 3: The image has no hate speech, slurs, or derogatory content. The tone is slightly sarcastic, criticizing excessive photo editing. 
Step 4: The context is lighthearted social media humor, not malicious or harmful. 
Step 5: Not offensive. \\[4pt]

\textbf{Metaphor} &
Step 1: Two children stand behind a table with a red chair built from objects such as a bucket, a drum, and a book. The background has a bookshelf filled with books. 
Step 2:“HONDA – The Power of Dreams.” 
Step 3: The creative chair symbolizes imagination and innovation. The children’s presence implies creativity is accessible to everyone. 
Step 4: The metaphor expresses that creativity and imagination can lead to achieving dreams, encouraging viewers to think outside the box. 
Step 5: Metaphorical. \\
\bottomrule
\end{tabular}

\label{tab:cot_examples}
\end{table}

\subsection{GRPO Prompt}
\label{sec:appendix_a2}
We use the prompt structure described by \citet{guo2025deepseek} for GRPO training as illustrated in Prompt \ref{prompt:grpo_sarcasm} to detect sarcasm. We apply the same GRPO structure to the other styles, which are omitted here since their prompts were introduced in the previous section.


\begin{prompt}[h]
    \centering
    \begin{promptboxstyle}
    You are an expert at detecting sarcasm in images and text. When given an image and text, analyze whether the content is sarcastic or not. Provide your reasoning process in the following format: Step 1: What the image shows: [Detailed description of the image content]. Step 2: What the caption says: [Quote or paraphrase the caption]. Step 3: Detecting mismatch: [Explain if there is a mismatch or congruence between the image and caption, and why]. Step 4: Inference of intent: [Conclude whether the intent is sarcastic or not based on the mismatch/congruence].
    Your reasoning process and answer should be enclosed within <think> </think> and <answer> </answer> tags, respectively. Answer with either sarcastic or not sarcastic in the answer tags, i.e.,<think> reasoning process (Step 1 to Step 4) </think><answer> sarcastic/not sarcastic </answer>        
    \end{promptboxstyle}
    \caption{GRPO prompt to detect sarcasm.}
    \label{prompt:grpo_sarcasm}
\end{prompt}

\clearpage
\section{Qualitative Examples}
In this section, we showcase outputs from our optimal model (SFT followed by GRPO on combined dataset). 
We include one representative example per style to highlight the model's reasoning as revealed by its generated CoT.
The sarcasm, humor, and metaphor examples reveal the model's ability to interpret image-text relationships. 
In sarcasm, the model identified consistency between the football team's image and the direct caption and accurately classified it as "Not Sarcastic." 
In humor, it recognized that combining incongruous images generated a comic effect. 
In metaphor, the model demonstrated abstract reasoning by detecting the symbolic link between the skier and the concept of sound. 
However, the offense example is notable because the model misclassified it, highlighting that defining offensive content is nuanced: While the model understood the joke about a character's nose, it erred on the side of caution and labeled it "Offensive" due to its reference to a physical trait. 
This example illustrates the challenge of aligning models with sensitive human judgments.

\tcbset{
  mybox1/.style={
    colframe=blue!70!black,
    colback=blue!5,
    arc=3pt, boxsep=2pt, outer arc=3pt,
    fonttitle=\bfseries, title filled,
    colbacktitle=blue!70!black
  },
  mybox2/.style={
    colframe=red!70!black,
    colback=red!5,
    arc=3pt, boxsep=2pt, outer arc=3pt,
    fonttitle=\bfseries, title filled,
    colbacktitle=red!70!black
  },
  mybox3/.style={
    colframe=green!70!black,
    colback=green!5,
    arc=3pt, boxsep=2pt, outer arc=3pt,
    fonttitle=\bfseries, title filled,
    colframe=green!60!black,
    colback=green!5, 
    coltitle=white,           
    colbacktitle=green!60!black
  }
}

\begin{tcolorbox}[
    title=Sarcasm Example, 
    mybox3,
    breakable,
    center title
]
    \begin{center}
        \includegraphics[width=0.5\linewidth]{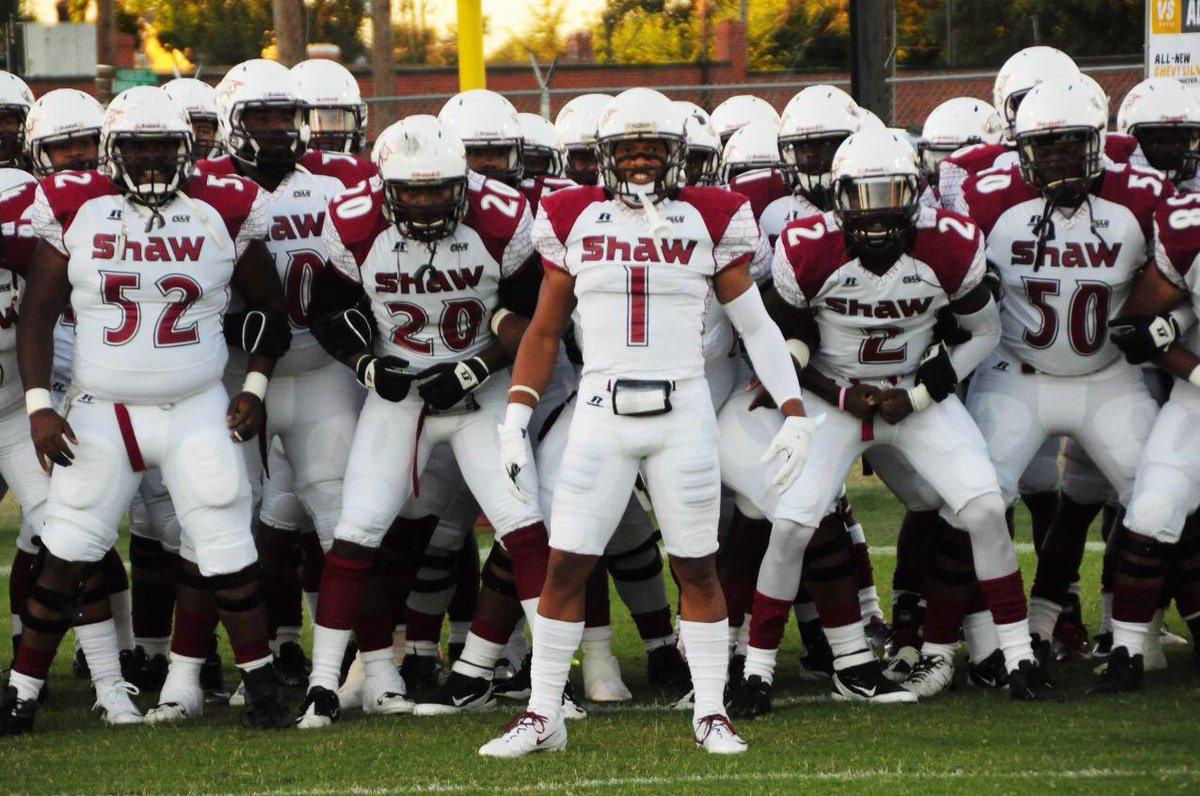}
    \end{center}
    \textbf{Text:} blessed to have received my first offer from shaw university \\
    \textbf{Ground Truth:} Not Sarcastic \\
    \textbf{CoT:} <think> Step 1: The image shows a group of football players in white uniforms with red accents, standing together on a field. They appear to be part of a team, possibly a college football team, given their uniforms and the setting. Step 2: The caption reads, \"Blessed to have received my first offer from Shaw University.\" This statement expresses gratitude for receiving an offer from Shaw University, which is likely a college or university where the player is interested in attending. Step 3: There is no apparent mismatch between the image and the caption. The image depicts a group of football players, and the caption directly relates to them, indicating that the player is expressing gratitude for receiving an offer from Shaw University. Step 4: Given that the caption is a direct statement about the player's feelings towards Shaw University, and there is no indication of sarcasm in the image or the caption, it can be inferred that the intent is not sarcastic.</think><answer>not sarcastic</answer>" \\
    \textbf{Prediction:} Not Sarcastic
\end{tcolorbox}

\begin{tcolorbox}[
    title=Humor Example, 
    mybox2,
    breakable,
    center title
]
    \begin{center}
        \includegraphics[width=0.5\linewidth]{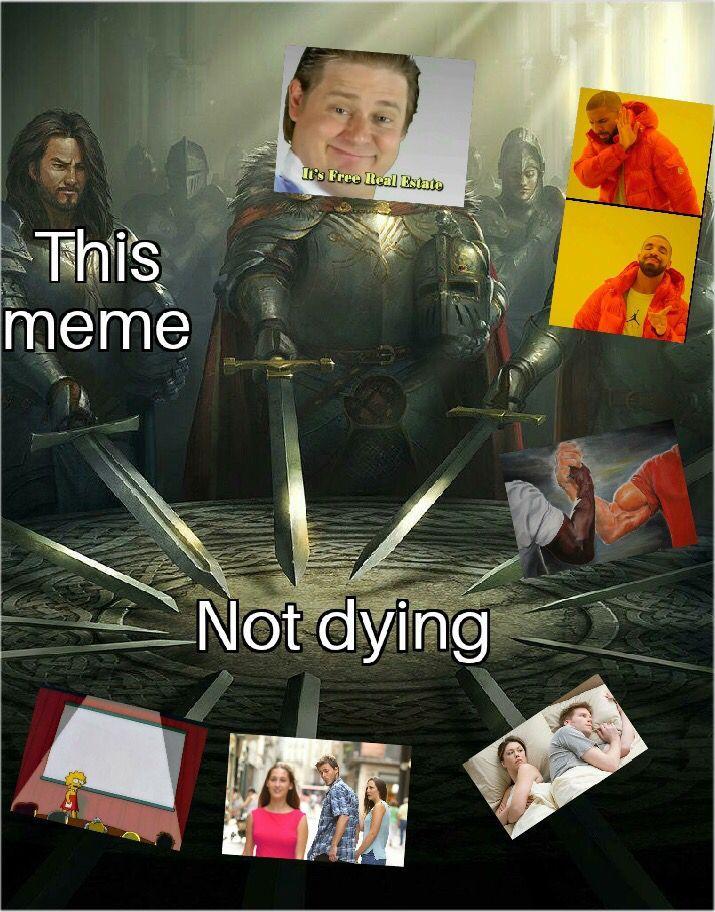}
    \end{center}
    \textbf{Text:} It's Free Real Estate This meme Not dying \\
    \textbf{Ground Truth:} Humorous \\
    \textbf{CoT:}  <think>  Step 1: The image is a collage of various memes, including a knight in armor holding a sword, a man with a caption, It's Free Real Estate, and several other images that appear to be unrelated to each other. The background of the image is dark and foggy, with a faint light shining down from above. Step 2: The caption reads ,It's Free Real Estate This meme Not dying. Step 3: The humor in this image comes from the juxtaposition of the serious-looking knight with the absurd and unrelated captions and images. The use of the phrase, It's Free Real Estate, as a caption for a knight holding a sword is unexpected and nonsensical, creating a sense of incongruity that is humorous. Additionally, the inclusion of other memes in the image, such as the man with the caption, It's Free Real Estate, adds to the absurdity and playfulness of the image. Step 4: Based on the humor cues present in the image, it appears that the intent of the creator is to be humorous. The use of absurd and unrelated captions and images creates a sense of playfulness and lightheartedness, suggesting that the creator is trying to make the viewer laugh.</think><answer> humorous </answer>", \\
    \textbf{Prediction:} Humorous
\end{tcolorbox}

\begin{tcolorbox}[title=Offense Example, 
    mybox1,
    breakable,
    center title
]
    \begin{center}
        \includegraphics[width=0.5\linewidth]{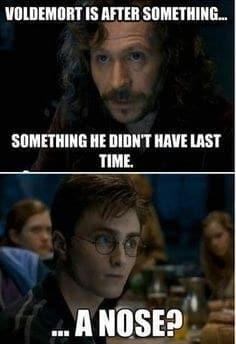}
    \end{center}
    \textbf{Text:} VOLDEMORT IS AFTER SOMETHING. SOMETHING HE DIDN'T HAVE LAST TIME. ...A NOSE? \\
    \textbf{Ground Truth:} Not offensive \\
    \textbf{CoT:} <think>Step 1: The image is a meme featuring two characters from the Harry Potter series, Voldemort and Harry Potter. Voldemort is depicted with a serious expression, while Harry Potter looks concerned. Step 2: The caption reads, VOLDEMORT IS AFTER SOMETHING... SOMETHING HE DIDN'T HAVE LAST TIME. ...A NOSE? Step 3: The caption contains a joke about Voldemort not having a nose, which could be perceived as making fun of individuals with physical disabilities or deformities. Step 4: The meme is likely intended to be humorous and lighthearted, rather than meant to offend or demean individuals with physical disabilities. However, the joke may still be perceived as insensitive or offensive by some viewers. </think><answer> offensive </answer>", \\
    \textbf{Prediction:} Offensive
\end{tcolorbox}

\begin{tcolorbox}[title=Metaphor Example,
    breakable,
    colframe=yellow!60!black,   
    colback=yellow!5,          
    coltitle=white,             
    colbacktitle=yellow!60!black, 
    center title
]
    \begin{center}
        \includegraphics[width=0.5\linewidth]{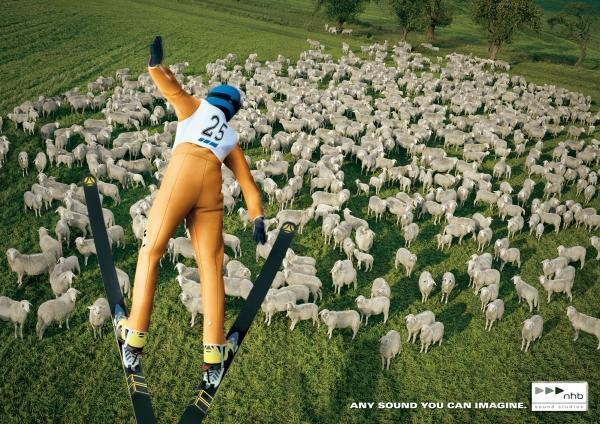} 
    \end{center}
    \textbf{Text:} any sound you can imagine \\
    \textbf{Ground Truth:} Metaphorical \\
    \textbf{CoT:}  <think>Step 1: The image depicts a person dressed in a yellow ski suit, complete with a blue helmet and black skis, standing in a field of sheep. The individual is positioned centrally in the frame, with their arms outstretched and their legs spread apart, as if they are about to take off or have just landed from a jump. The surrounding environment is characterized by a lush green field, dotted with numerous white sheep grazing peacefully. Step 2: The caption reads, any sound you can imagine. This phrase is concise and enigmatic, leaving room for interpretation. Step 3: Upon closer examination, it becomes apparent that the image and caption are connected through metaphorical language. The juxtaposition of the skier amidst the sheep, coupled with the caption's reference to sound, suggests a symbolic comparison between the skier's actions and the auditory experience. The use of the word imagine further reinforces this interpretation, implying that the viewer should consider the skier's movements as a form of sound or music. Step 4: The metaphorical connection between the image and caption conveys the idea that the skier's actions can be perceived as a form of creative expression, akin to music or sound. The caption's emphasis on imagination encourages the viewer to consider the skier's movements as a means of generating sound or music, thereby blurring the lines between physical activity and auditory experience. </think><answer> metaphorical </answer>" \\
    \textbf{Prediction:} Metaphorical

\end{tcolorbox}



\end{document}